\documentclass{article}

\PassOptionsToPackage{numbers, compress}{natbib}
\usepackage[preprint]{neurips_2026}

\usepackage[utf8]{inputenc} 
\usepackage[T1]{fontenc} 
\usepackage{hyperref} 
\usepackage{url} 
\usepackage{booktabs} 
\usepackage{amsfonts} 
\usepackage{nicefrac} 
\usepackage[nopatch=footnote]{microtype} 
\usepackage{xcolor} 
\usepackage{graphicx} 
\usepackage{amsmath} 
\usepackage{multirow}
\usepackage{wrapfig}
\usepackage{float}
\usepackage{calc}
\usepackage{caption}

\title{High Dynamic Range 3D Gaussian Splatting via Luminance–Chromaticity
Decomposition}

\author{%
  \textbf{Kaixuan Zhang}$^1$
  \quad
  \textbf{Minxian Li}$^1$\thanks{Minxian Li (minxianli@njust.edu.cn) is the corresponding author with School of Computer Science and Engineering, Nanjing University of Science and Technology.}
  \quad
  \textbf{Mingwu Ren}$^1$ \\ 
  \quad
  \textbf{Jiankang Deng}$^2$
  \quad
  \textbf{Xiatian Zhu}$^3$
  \vspace{.5em} 
  \\
  $^1$Nanjing University of Science and Technology
  \qquad
  $^2$Imperial College London \\ 
  $^3$University of Surrey
}

\begin{document}
\maketitle

\begin{abstract}
High Dynamic Range (HDR) 3D reconstruction is pivotal for professional content creation in filmmaking and virtual production. Existing methods typically rely on multi-exposure Low Dynamic Range (LDR) supervision to constrain the learning process within vast brightness spaces, resulting in complex, dual-branch architectures. This work explores the feasibility of learning HDR 3D models exclusively in the HDR data space to simplify model design. By analyzing 3D Gaussian Splatting (3DGS) for HDR imagery, we reveal that its failure stems from the limited capacity of Spherical Harmonics (SHs) to capture extreme radiance variations across views, often biasing towards high-radiance observations and underfitting. While increasing the maximum SH degree improves training fitting, it leads to severe overfitting and excessive parameter overhead. To address this, we propose \textit{Luminance--Chromaticity Decomposition Gaussian Splatting} (LCD-GS). By decoupling luminance and chromaticity into independent parameters, LCD-GS significantly enhances learning flexibility with minimal parameter increase (\textit{e.g.}, one extra scalar per primitive). Notably, LCD-GS maintains the original training and inference pipeline, requiring only a change in color representation. 
This explicit decomposition naturally enables primitive-level local and global luminance editing during inference.
Extensive experiments on synthetic and real datasets demonstrate that LCD-GS consistently outperforms state-of-the-art methods in reconstruction fidelity and dynamic-range preservation even with a simpler, more efficient architecture, providing an elegant paradigm for professional-grade HDR 3D modeling. Code and datasets will be released.
\end{abstract}

\section{Introduction}
\label{sec:intro}

\begin{wrapfigure}{r}{0.32\textwidth}
\centering
\vspace{-5.0em}
\includegraphics[width=\linewidth]{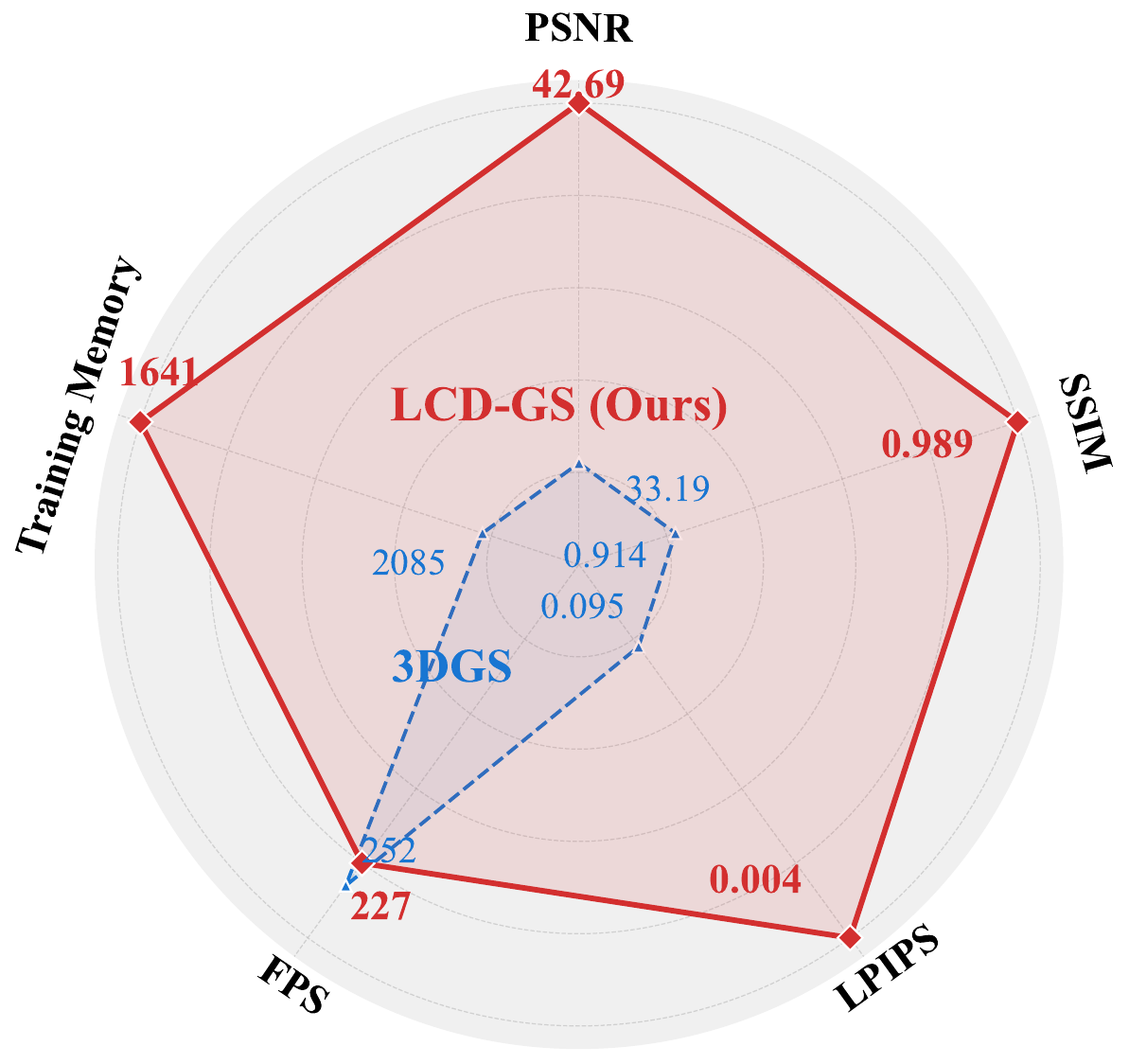}
\vspace{-2.0em}
\caption{LCD-GS unlocks 3DGS in HDR 3D modeling while keeping the efficiency.
}
\label{fig:radar_chart}
\vspace{-1.5em}
\end{wrapfigure}
\begin{figure}[tbp]
\centering
\includegraphics[width=\linewidth]{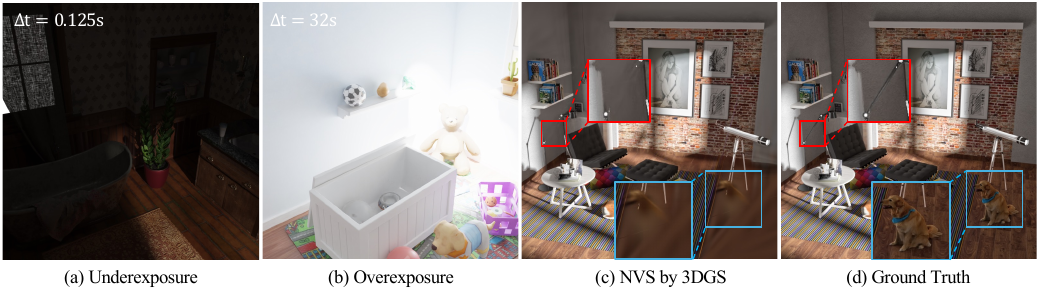}
\vspace{-1em}
\caption{Example of HDR 3D modeling: (a) underexposure and (b) overexposure. (c) 3DGS \cite{3dgs} exhibits severe blurring artifacts. (d) ground truth. $\Delta t$: Exposure time.}
\label{fig:motivation}
\vspace{-2.0em}
\end{figure}

3D scene modeling such as Novel View Synthesis (NVS) \cite{reddy2025survey, cai2024nerf} has achieved remarkable progress, enabling photorealistic rendering for gaming \citep{pan2024game}, AR/VR \citep{van2022novel}, and autonomous driving \citep{ma2025novel}. Existing NVS models \citep{3dgs,dalal2024gaussian,niu2024overview} are typically confined to LDR imagery, suffering from severe under- and over-exposure artifacts (see Fig.~\ref{fig:motivation}a-b) that prevent the recovery of accurate physical lighting. To alleviate this and reconstruct full dynamic range, High Dynamic Range Novel View Synthesis (HDR NVS) methods \citep{hdrnerf,hdr-gs,gausshdr,instanthdr} have been built often leveraging multi-exposure LDR supervision. This paradigm is designed to constrain the learning process within vast brightness spaces, but consequently necessitates complex, dual-branch architectures with specialized inverse tone-mapping or fusion pipelines. Such designs significantly increase capture complexity, require aligned exposure stacks, and are highly sensitive to motion misalignment \citep{kalantari2017deep}.

To overcome these structural limitations and simplify model design, here we propose to explore \textit{the feasibility of learning HDR 3D models exclusively in the HDR data space}. The growing integration of HDR-oriented imaging pipelines in modern devices (\textit{e.g.}, iPhones, Sony IMX490) makes acquiring such richer, direct radiometric observations increasingly accessible \citep{wang2016real}, bypassing the need for tedious multi-exposure setups. However, our analysis reveals that applying 3DGS \citep{3dgs}, a state-of-the-art 3D model\footnote{NeRF \cite{nerf} encounters similar representational limitations due to the entanglement of geometric density and radiance within a single neural network (see Tab.~\ref{tab:res_syn}).}, directly to HDR data leads to catastrophic failure: the resulting radiance fields become heavily biased towards bright regions and fail to learn darker, low-exposure areas (see Fig.~\ref{fig:motivation}c).

\begin{figure*}[tbp]
\centering
\includegraphics[width=\linewidth]{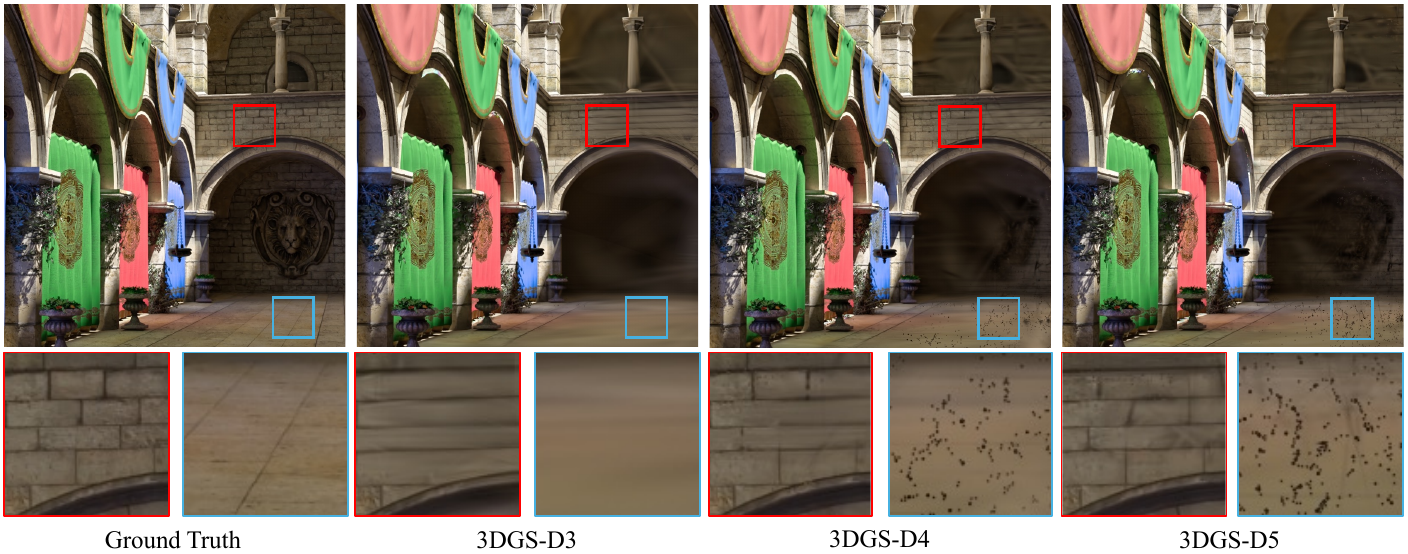}
\caption{Elevating the maximum SH degree ($D$) of 3DGS mitigates blurring artifacts while simultaneously causing additional artifacts. D$x$ means the maximum SH degree is $x$, 3 by default.}
\label{fig:sh3vs4}
\vspace{-1.0em}
\end{figure*}

We conjecture this failure fundamentally stems from \textit{the limited flexibility of 3DGS's color representation}. Concretely, its Spherical Harmonics (SHs) based formulation \textit{entangles} luminance and chromaticity into \textit{a unified set of coefficients} (see Fig. \ref{fig:3d-sh}), which can encode a modest range of colors (\textit{e.g.}, 256 values per channel in 8-bit case) sufficient for typical LDR image rendering. Yet, when applied to HDR observations (\textit{e.g.}, 12/32-bit per channel) where radiance can vary by orders of magnitude exponentially across viewing directions, 3DGS fails to capture sharp, drastic color variations, leading to substantial underfitting (see Fig. \ref{fig:intensity_curves}). Finally, a trained 3DGS model would be biased to prioritize high-radiance observations as they incur bigger errors and dominate the optimization. We further investigate whether increasing the maximum SH degree of 3DGS's color representation (\textit{e.g.}, from the default degree 3 to 4/5) helps. While elevating the maximum SH degree marginally improves the fitting of training data, it ultimately leads to severe overfitting instead, introduces additional artifacts (see Fig. \ref{fig:sh3vs4} and Tab. \ref{tab:res_syn}), and incurs excessive parameter overhead.

To address this inherent limitation without inducing prohibitive costs, we propose \textit{Luminance-Chromaticity Decomposition Gaussian Splatting} (LCD-GS), inspired by the above insights, as well as human visual system's distinct functional pathways for processing luminance and chromaticity \citep{Xing2226}, LCD-GS explicitly \textit{decouples luminance and chromaticity} in parameterization. We factorize the radiance of each 3D Gaussian primitive into a \textit{view-independent positive luminance scalar}, which represents absolute radiance magnitude, and a \textit{view-dependent chromaticity term}, which represents an objective specification of the color (regardless of its luminance). This disentanglement significantly enhances learning flexibility by offering full freedom to both the luminance and chromaticity. The chromaticity term can be represented with SH same as 3DGS's color. Beyond improved representational learning flexibility, this decoupling strategy further enables controllable luminance editing of scenes (\textit{e.g.}, highlight suppression and shadow enhancement in Appendix \ref{sec:app:scene_edition}). Notably, LCD-GS preserves the original training and inference pipeline of 3DGS, requiring only a change in color representation and a minimal parameter increase (\textit{e.g.}, one scalar for luminance per primitive).

Our \textbf{contributions} are: \textbf{(I)} We present the first systematic study of learning HDR 3D models \textit{exclusively} in the HDR RGB data space, opening up a simplified, more practical paradigm for professional digital content creation. \textbf{(II)} We reveal the fundamental limitations of existing 3D models in processing HDR observations along with insightful analysis and discussions. \textbf{(III)} We propose LCD-GS, a novel 3DGS variant characterized by luminance--chromaticity decomposition for more scalable radiance representation and learning from HDR imagery. 
\textbf{(IV)} Extensive evaluations demonstrate that, even with simplified architecture and no LDR supervision, LCD-GS outperforms all alternatives on both synthetic and real-world benchmarks, validating its radiometric accuracy and efficiency (see Fig. \ref{fig:radar_chart}).

\section{Related Work}
\label{sec:related_work}
\noindent\textbf{Novel view synthesis (NVS)} is a fundamental task in 3D vision, with applications in AR/VR \citep{van2022novel}, gaming \citep{pan2024game}, and autonomous driving \citep{ma2025novel}. Classical methods like Structure-from-Motion (SfM) \citep{pan2024global} and Multi-View Stereo (MVS) \citep{huang2025multi} rely on geometric cues from multi-view images but struggle with occlusions, textureless regions, and high computational costs \citep{feng2025survey}. Recent learning-based approaches model scenes as continuous differentiable representations. NeRF \citep{nerf} pioneered this paradigm by using neural networks to map 3D coordinates and viewing directions to volume density and color via differentiable volume rendering. Subsequent works \citep{gao2024mip, zhu2024spikenerf, yao2025spinerf, mueller2022instant, yu2021plenoctrees,Fu_2024_CVPR} aim to improve efficiency and quality. Alternatively, 3DGS \citep{3dgs} and its variations \citep{absgs,Yu_2024_CVPR,cheng2024gaussianpro,kulhanek2024wildgaussians} represent scenes as learnable 3D Gaussian primitives, bypassing volumetric rendering and enabling real-time performance. Despite progress, such existing state-of-the-art NVS methods are predominantly focused on LDR sRGB inputs, limiting their ability to reconstruct or render true HDR radiance from real-world scenes.

\noindent\textbf{HDR novel view synthesis (HDR NVS)} aims to reconstruct scenes with high dynamic range from multi-view observations. Huang \textit{et al.} \citep{hdrnerf} introduced HDR-NeRF, the first HDR-NVS framework, which extends standard NeRF \citep{nerf} to learn mappings from physical radiance to HDR color using multi-exposure LDR inputs. However, its reliance on the NeRF architecture results in prohibitively slow inference. HDR-GS \cite{hdr-gs} addressed this limitation using 3DGS \citep{3dgs}, achieving significantly faster rendering and improved visual quality. GaussHDR \cite{gausshdr} further introduced a unified tone-mapping strategy that enables HDR novel view synthesis without requiring HDR supervision. Despite these advances, all aforementioned methods fundamentally rely on multi-exposure LDR image stacks, which entail complex capture protocols and substantial storage overhead.

The increasing availability of HDR-oriented imaging pipelines and HDR-capable sensors (\textit{e.g.}, Sony IMX490) motivates our study \citep{wang2016real}. Compared with standard LDR imagery, such data provide richer radiometric information, making them attractive for HDR-aware 3D reconstruction. While prior HDR-aware methods such as RawNeRF \citep{rawnerf}, HDRSplat \citep{hdrsplat}, and LE3D \citep{le3d} reconstruct HDR scenes from linear RAW observations, typically under noisy or low-light conditions, we instead target direct HDR reconstruction from RGB-domain HDR images. This setting removes the dependence on RAW sensor data and introduces a distinct representation challenge: how to faithfully model extreme radiance variation directly in HDR RGB space. To the best of our knowledge, we present the first study to directly operate on HDR RGB imagery without relying on LDR-based supervision. Furthermore, by introducing a principled luminance--chromaticity decomposition, LCD-GS directly models extreme radiance variations, enabling highly faithful HDR scene reconstruction and luminance editing. Further discussion of related illumination editing is provided in the Appendix \ref{sec:app:add_related}.

\section{Method}
\label{sec:method}

\textbf{Problem setting.} We aim to learn an HDR 3D model $\mathcal{F}$ for a target scene from posed HDR images, $\mathcal{F}:(V) \rightarrow \mathbf{I}_{V}$, that can render an HDR image $\mathbf{I}_{V}$ for any viewpoint $V$. To that end, we capture a set of HDR training images $\mathbf{I}=\{\mathbf{I}_{1}, \cdots, \mathbf{I}_{q}, \cdots, \mathbf{I}_{Q}\}$, with $\mathbf{I}_{q}$ the $q$-th image taken under view $V_{q}$.


\subsection{Preliminary: color representation in 3DGS}
\label{sec:method:preliminary}
In vanilla 3DGS \citep{3dgs}, given a pixel located at $\boldsymbol{x}$ and its distances to all overlapping Gaussian primitives along the viewing ray, alpha compositing is employed to compute the final pixel color:
\begin{equation}
\boldsymbol{p(x)}=\sum_{n=1}^{N}\boldsymbol{c}_{n}\alpha'_{n}\prod_{j=1}^{n-1}\left(1-\alpha'_{j}\right),
\end{equation}
where $N$ denotes the number of Gaussian primitives intersected by the ray in a front-to-back order. Here, $\boldsymbol{c}_{n}\in \mathbb{R}^{3}$ represents the color of the $n$-th primitive, $\alpha'_{n}\in [0,1]$ is its effective opacity at pixel $\boldsymbol{x}$ (obtained from the projected 2D Gaussian and its density), and $\prod_{j=1}^{n-1}(1-\alpha'_{j})$ denotes the accumulated transmittance accounting for occlusion by all preceding Gaussian primitives along the ray. The resulting $\boldsymbol{p(x)} \in \mathbb{R}^{3}$ is the rendered RGB color of the pixel.

The color of each Gaussian primitive is modeled as a view-dependent function using spherical harmonics (SHs, see Fig. \ref{fig:3d-sh} in the Appendix). Specifically, each Gaussian primitive is associated with a set of SH coefficients $\mathbf{k}=\{\boldsymbol{k}_{d}^{m}|0\le d\le D, -d\le m\le d\}\in \mathbb{R}^{(D+1)^2\times 3}$ that model view-dependent appearance, and each $\boldsymbol{k}_{d}^{m}\in \mathbb{R}^{3}$ is a set of three coefficients corresponding to the RGB components where the index $m$ denotes the azimuthal order of the SH basis function $Y_{d}^{m}$, controlling its angular frequency and phase around the polar axis for a given degree $d$. $D$ denotes the maximum SH degree, by default $D=3$. Formally, the color of the $n$-th Gaussian primitive under view $\boldsymbol{v}=(\theta,\phi)$ is expressed as
\begin{equation}
\label{eq:3dgs}\boldsymbol{c}_{n}(\boldsymbol{v}|\mathbf{k}) = \sum_{d=0}^{D}\sum_{m=-d}^{d}\boldsymbol
{k}_{d}^{m}Y_{d}^{m}(\theta,\phi),
\end{equation} 
where $Y_{d}^{m}: \mathbb{S}^{2}\rightarrow \mathbb{R}$ is the SH function that maps 3D points on the sphere to real numbers \cite{3dgs,hdr-gs}.

\noindent {\bf Remarks.} The color formulation of 3DGS (\textit{i.e.}, Eq. \eqref{eq:3dgs}) implicitly assumes that radiance variations of input images
 across views are moderate \citep{wang2006all}. Under such cases, the SH expansion provides a compact low-frequency approximation of the bidirectional color variation. 
However, when applied to HDR imagery, where radiance can vary by \textit{orders of magnitude} across different views, this color representation becomes particularly susceptible, struggles to capture sharp, drastic color variations, leading to underfitting (see Fig. \ref{fig:intensity_curves} and Tab. \ref{tab:res_syn}).

\subsection{Baseline: Elevating the maximum SH degree in 3DGS}
\label{sec:method:elevating} 
The above critique raises a natural question: \textit{Can increasing the maximum SH degree alleviate these limitations}? An intuitive approach is to elevate the maximum SH degree in 3DGS \citep{3dgs}, \textit{e.g.}, increasing from the default maximum degree of 3 (16 coefficients) to 5 (36 coefficients, see Fig.~\ref{fig:app:shs_vis} in the Appendix). We find that although increasing the maximum SH degree marginally improves the quantitative metrics (see rows 2--4 of Tab. \ref{tab:res_syn}), they fail to enhance perceptual rendering quality (see Fig. \ref{fig:sh3vs4}), while slowing down. This indicates fundamental limitations of 3DGS's color formulation for HDR modeling, going beyond the maximum SH degree.

\begin{figure}
\centering
\includegraphics[width=\linewidth]{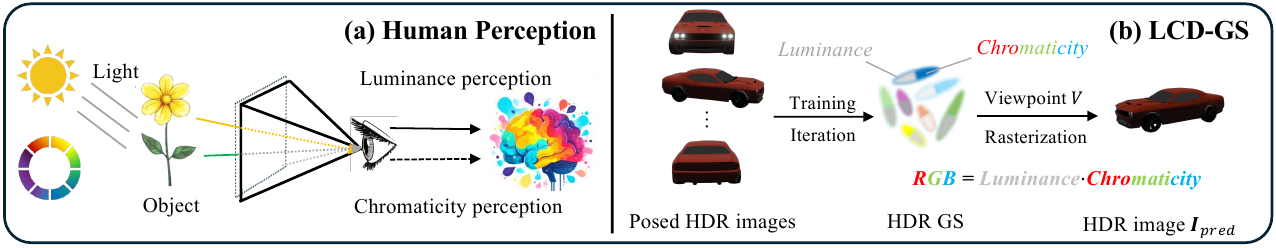}
\caption{\textit{Overview of LCD-GS}. With posed HDR images of a scene, we aim to learn an HDR Gaussian Splatting model. We draw inspiration from \textbf{(a)} human visual perception on tackling drastic radiance variation: luminance and chromaticity are perceived \textit{separately}. In analogy, \textbf{(b)} LCD-GS decomposes the parameterization of luminance and chromaticity for color representation, in contrast to the conventional SH color representation of 3DGS which tightly entangles the two aspects using a single set of parameters. Concretely, LCD-GS computes the color via multiplicative composition of luminance and chromaticity, each of which can be freely optimized for arbitrary combinations.}
\label{fig:overview}
\vspace{-1.5em}
\end{figure}

\subsection{Proposed: LCD-GS}
\label{sec:method:ours} 
To address the above limitations, we propose LCD-GS, illustrated in Fig.~\ref{fig:overview}, which introduces a luminance--chromaticity decomposition for the color modeling of Gaussian primitives. Instead of directly representing the color of Gaussian primitives using RGB spherical harmonics, we decompose the radiance of each Gaussian primitive into a view-independent positive luminance scalar and a view-dependent bounded chromaticity term. Formally, the color emitted by the $n$-th Gaussian primitive along the viewing direction $\boldsymbol{v}=(\theta,\phi)$ is defined as:
\begin{equation}
\label{eq:ours}\boldsymbol{c}_{n}(\boldsymbol{v}|\mathbf{k}, L) = \underbrace{L
\vphantom{\mathcal{N} \left( \sum_{d=0}^D \sum_{m=-d}^d \boldsymbol{k}_d^m Y_d^m(\theta,\phi)
\right)}}_{\text{Luminance}}\cdot \underbrace{\mathcal{N} \left(\left[\sum_{d=0}^D
\sum_{m=-d}^d \boldsymbol{k}_d^m Y_d^m(\theta,\phi)\right]_+ \right)}_{\text{Chromaticity}},
\end{equation}
where the luminance term $L \in \mathbb{R}^{+}$ is a view-independent positive scalar controlling the overall radiance magnitude. To enforce non-negativity, we parameterize it as $L=g(l)$, where $l$ is learnable and $g:\mathbb{R}\rightarrow\mathbb{R}^{+}$ is a positive activation function, such as $\mathrm{softplus}(l)=\log(1+e^{l})$ (see Appendix~\ref{sec:app:ablation} for the ablation study).
The chromaticity term is the normalized SH-based component, which captures view-dependent color variation in a bounded color space. 
Here, $\boldsymbol{k}_d^m$ denotes the learnable SH coefficient of degree $d$ and order $m$, and $Y_d^m(\theta,\phi)$ is the corresponding SH basis evaluated along the viewing direction $\boldsymbol{v}=(\theta, \phi)$ (see Sec. \ref{sec:method:preliminary}). 
The operator $[\cdot]_+=\max(\cdot,0)$ is applied channel-wise to ensure non-negative chromatic responses, while $\mathcal{N}(\cdot)$ denotes $L_1$ normalization, producing a scale-invariant color representation per primitive.

\noindent\textbf{Discussion.}
The proposed luminance--chromaticity decomposition offers several advantages.
\textit{First}, it explicitly decouples view-independent radiance scale from view-dependent chromatic variation, which are tightly entangled in the original SH color representation of vanilla 3DGS. This makes the formulation particularly suitable for HDR representation: the luminance component can accommodate a wide dynamic range through an unbounded positive scalar parameterization, while the chromaticity component remains in a normalized, stable, and physically plausible color space.
\textit{Second}, the formulation can be easily integrated into exisiting 3DGS methods \citep{3dgs, Fu_2024_CVPR, absgs, Yu_2024_CVPR} by replacing only the color parameterization in Eq. \eqref{eq:3dgs} with Eq. \eqref{eq:ours}, while keeping the rest of the pipeline unchanged.
\textit{Third}, the explicit luminance component provides a simple and interpretable handle for luminance editing: one can manipulate the learned $L$ globally or locally while keeping chromaticity, opacity, and geometry unchanged, enabling basic exposure adjustment and HDR radiance editing without introduce extra networks \citep{liang2024gs, shi2025gir} or retraining (see Sec. \ref{sec:method:scene_edition}).

\subsection{Model optimization}
To mitigate optimization instabilities caused by the high dynamic range of radiance values, we apply a $\mu$-law compression function $\mathcal{T}_{\mu}(\boldsymbol{p}) = \frac{\log(1+\mu\cdot \boldsymbol{p})}{\log(1+\mu)}$ during training,
where $\boldsymbol{p}$ denotes a pixel RGB value in HDR and $\mu\in R^+$ controls the compression strength, following established works \cite{hdrnerf, hdr-gs, mono-hdr-3d}.
Following this transformation, the overall training objective is defined as:
\begin{equation}
\label{eq:loss}\mathcal{L}= \lambda \cdot \mathcal{L}_{1}(\hat{\mathbf{I}}_{\text{pred}}
, \hat{\mathbf{I}}_{\text{gt}}) + (1-\lambda) \cdot \mathcal{L}_{SSIM}(\hat
{\mathbf{I}}_{\text{pred}}, \hat{\mathbf{I}}_{\text{gt}}),
\end{equation}
where $\mathbf{\hat{I}}_{\text{pred}}$ and $\mathbf{\hat{I}}_{\text{gt}}\in \mathbb{R}^{H\times W\times C}$ represent rendered and HDR ground truth, respectively, both transformed by the $\mu$-law compression, and $\lambda$ controls the trade-off between the two loss terms.

\begin{figure}[tbp]
\vspace{-1.5em}
\centering
\includegraphics[width=\linewidth]{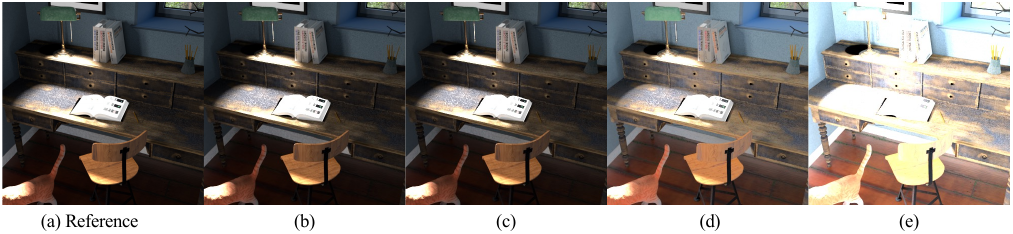}
\caption{Local luminance editing enabled by LCD-GS. (a) Reference image. (b)-(e) Shadow enhancement by scaling the luminance of a subset of low-luminance Gaussian primitives.}
\label{fig:app:lum_edit_shadow}
\vspace{-1.8em}
\end{figure}

\subsection{Luminance editing}
\label{sec:method:scene_edition} 
A unique advantage of our proposed LCD-GS is its inherent capability for luminance editing. As detailed in Sec. \ref{sec:method:ours}, LCD-GS explicitly decouples the radiance of each 3D Gaussian primitive into a view-independent luminance scalar $L$ and a normalized chromaticity term. Since the chromaticity information is completely isolated from the radiance magnitude, we can effortlessly manipulate the scene's luminance without destroying the underlying colors or introducing hue shifts.
Specifically, luminance editing can be performed by applying a user-defined mapping function $f:\mathbb{R}^{+}\rightarrow\mathbb{R}^{+}$ to the learned luminance values of all Gaussian primitives, or to a selected subset for local editing. During rendering, the original luminance $L$ is replaced with $L' = f(L)$, while the chromaticity term remains unchanged. As shown in Fig. \ref{fig:app:lum_edit_shadow}, this enables local enhancement of shadowed regions by adjusting low-luminance Gaussian primitives whose luminance values fall below a predefined threshold. Additional details and discussions are provided in Appendix \ref{sec:app:scene_edition}.

\vspace{-0.5em}
\section{Experiments}
\label{sec:experiments} \textbf{Datasets.} To validate the performance of LCD-GS, we conduct extensive experiments on both synthetic and real-world data. We
adopt the \text{synthetic} HDR dataset from HDR-NeRF \cite{hdrnerf}, comprising 8 scenes rendered at $800\times 800$ resolution in Blender \cite{blender},
named as \textbf{\textit{HDR-Syn-8S}}. Each scene contains 35 multi-view HDR images, with every HDR image explicitly paired with 5 corresponding LDR captures
taken at distinct shutter speeds. Critically, this multi-exposure LDR acquisition protocol which demands precise exposure bracketing, vibration-free camera rigs, and per-scene recalibration represents a costly and tedious process that severely limits real-world deployment scalability.

For real-world evaluation, we introduce \textbf{\textit{iHDR-4S}}, a new multi-view HDR dataset comprising 4 indoor scenes with geometrically and photometrically complex characteristics captured by an iPhone 14 Pro. Unlike existing datasets that rely on synthetic HDR data or simple exposure bracketing, iHDR-4S provides HDR captures with precise camera poses, making it a valuable resource for HDR-aware 3D reconstruction research. Please refer to Sec. \ref{sec:app:datasets} in the Appendix for more details.

\noindent\textbf{Implementation details.} LCD-GS is built upon PyTorch \cite{pytorch} and trained with Adam optimizer. The learning rate of luminance attribute is set to 0.05, and the activation function $g$ for luminance is the softplus function. All the experiments are conducted with a single NVIDIA RTX 4090 GPU. $\lambda$ and $\mu$ are set to 0.2 and 5000 during all the experiments. When adapting NeRF \cite{nerf} for HDR image training on the synthetic dataset, we replace its output activation function from sigmoid to softplus, enabling the model to predict unbounded radiance values while preserving the original architecture and training pipeline. If necessary, compared methods \cite{hdr-gs, gausshdr, mono-hdr-3d} are granted access to LDR inputs on the HDR-Syn-8S dataset per their original implementations. For all experiments, we uniformly downsample the input images by a factor of 2 to reduce computational cost.

\noindent\textbf{Evaluation metrics.} We use PSNR and SSIM as the primary evaluation metrics, and additionally report LPIPS to assess perceptual similarity. For efficiency analysis, we also report training time, peak training memory consumption, and inference speed in frames per second (fps). {\em Results are averaged over all scenes}. Following established practice in prior works \cite{hdrnerf, hdr-gs, gausshdr, mono-hdr-3d}, we also quantitatively evaluate the rendered HDR images in the tone-mapped domain and qualitatively show HDR results tone-mapped by Photomatix Pro \cite{Photomatix_pro}.

\begin{table}[tb]
\begin{minipage}{\textwidth}
\caption{Results on the HDR-Syn-8S dataset. 
The \textbf{best} and \underline{second-best} results are highlighted.}
\centering
\renewcommand{\arraystretch}{1.0}
\setlength{\tabcolsep}{2pt}
\small \resizebox{\textwidth}{!}{
\begin{tabular}{c|l|c|cc|ccc|c}
\toprule[0.15em] \multirow{2}{*}{Row} & \multirow{2}{*}{Method}  & \multirow{2}{*}{LDR Supervision} & \multicolumn{2}{c|}{Training} & \multicolumn{3}{c|}{HDR} & Inference          \\
&     &       & Time (min)   & Memory (MB)     & PSNR$\uparrow$  & SSIM$\uparrow$ & LPIPS$\downarrow$ & Speed (fps)     \\
\midrule[0.1em] 
1   & NeRF \citep{nerf}     & $\times$     & 264          & 11575  & 15.20   & 0.388    & 0.753    & 0.42      \\
2   & 3DGS \citep{3dgs}     & $\times$     & \textbf{9}   & \underline{2085}  & 33.19   & 0.914    & 0.095    & \textbf{252}    \\
3   & 3DGS-D4   & $\times$   & \underline{10}  & 2518  & 34.93  & 0.839  & 0.075  & 205   \\
4   & 3DGS-D5   & $\times$   & \underline{10} & 2616 & 35.37  & 0.948  & 0.067    & 148   \\
5   & HDR-NeRF \citep{hdrnerf}  & \checkmark  & 543  & 12373  & 36.40  & 0.936   & 0.018  & 0.12  \\
6   & HDR-GS \citep{hdr-gs}  & \checkmark  & 21  & 3761  & 38.29  & 0.968  & 0.014   & 126  \\
7   & Mono-HDR-GS \citep{mono-hdr-3d}   & \checkmark   & 44   & 6329  & 38.57  & 0.970  & 0.013  & 137  \\
8   & Mono-HDR-NeRF \citep{mono-hdr-3d} & \checkmark   & 439  & 9281  & 32.86  & 0.940  & 0.068  & 0.26   \\
9   & GaussHDR \citep{gausshdr}         & \checkmark   & 31   & 5277  & \underline{39.02}  & \underline{0.976}  & \underline{0.010}  & 23  \\
10  & \textbf{LCD-GS (Ours)} & $\times$ & 14  & \textbf{1641} & \textbf{42.69} & \textbf{0.989} & \textbf{0.004} & \underline{227} \\
\bottomrule[0.15em]
\end{tabular}
} 
\label{tab:res_syn}
\end{minipage}


\begin{minipage}{\textwidth}
\centering
\includegraphics[width=1.0\linewidth]{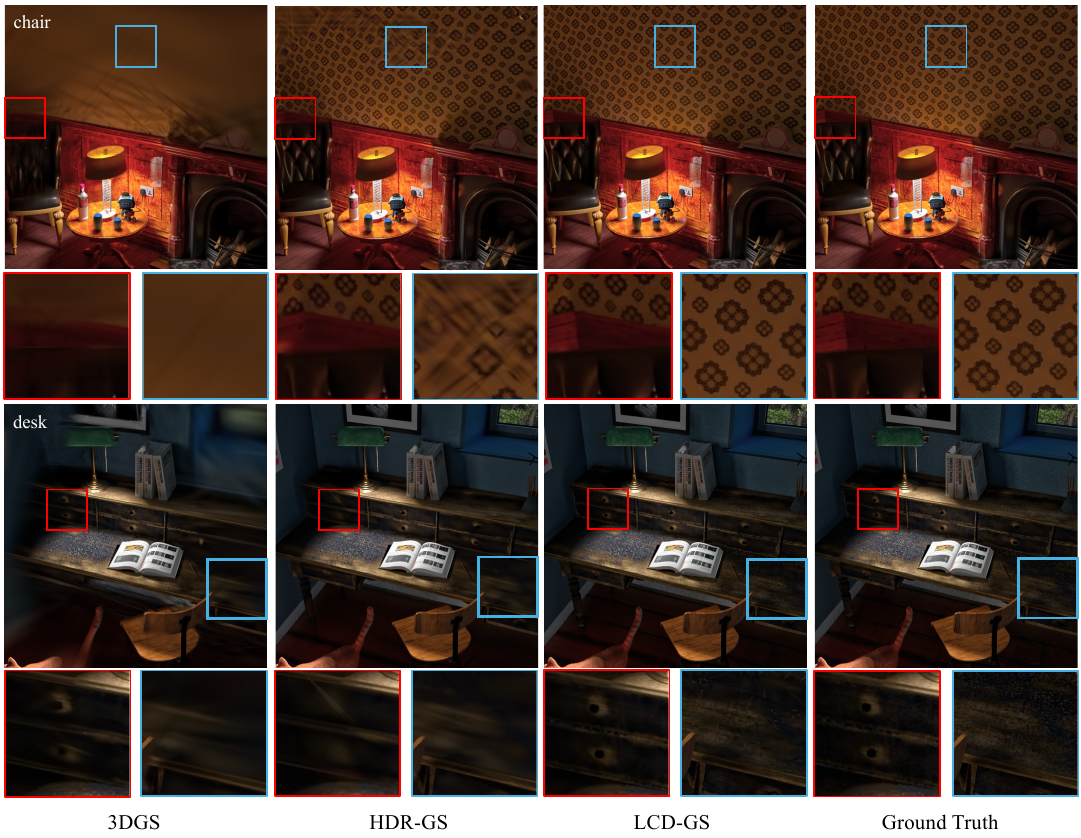}
\vspace{-1.0em}
\captionof{figure}{Visual comparisons on the HDR-Syn-8S dataset.}
\label{fig:syn_cmp}
\vspace{-2.0em}
\end{minipage}
\end{table}

\subsection{Quantitative evaluation}
\textbf{Competitors.} On the synthetic HDR-Syn-8S dataset, we compare LCD-GS with both LDR-focused and HDR-specific NVS methods. For LDR-focused baselines, we include NeRF \cite{nerf} and several 3DGS \cite{3dgs} variants using SHs with different maximum degrees. These baselines serve to examine whether simply extending conventional LDR-oriented NVS methods to HDR inputs is sufficient for HDR scene reconstruction. We further compare against five representative HDR NVS methods: (1) HDR-NeRF \cite{hdrnerf}, a pioneering NeRF-based method for synthesizing HDR novel views from multi-exposure LDR inputs; (2) HDR-GS~\cite{hdr-gs}, which extends the efficient explicit representation of 3DGS to HDR radiance modeling using multi-exposure LDR images and HDR supervision; (3)--(4) Mono-HDR-GS and Mono-HDR-NeRF~\cite{mono-hdr-3d}, which aim to reconstruct HDR scenes from single-exposure LDR inputs with HDR supervision; and (5) GaussHDR~\citep{gausshdr}, a recent Gaussian-splatting-based HDR NVS method designed to improve HDR reconstruction quality through enhanced radiance modeling from multi-exposure LDR images.

On the real-world iHDR-4S dataset, we compare LCD-GS with NeRF and 3DGS variants, as well as Mono-HDR-GS and Mono-HDR-NeRF. Since iHDR-4S contains native HDR images but does not provide aligned multi-exposure LDR stacks, methods that require multi-exposure LDR inputs, including HDR-NeRF, HDR-GS, and GaussHDR, are not applicable under their original settings. For Mono-HDR-GS and Mono-HDR-NeRF, we convert the HDR images in iHDR-4S into single-exposure LDR counterparts using a simple ISP pipeline, enabling their training under the required single-exposure LDR input setting.

\begin{figure}[tb]
\centering

\begin{minipage}{\linewidth}
\centering
\captionof{table}{Results on the iHDR-4S dataset. 
The \textbf{best} and \underline{second-best} results are highlighted.}
\renewcommand{\arraystretch}{1.0}
\resizebox{\textwidth}{!}{
\begin{tabular}{c|l|c|cc|ccc|c}
\toprule[0.15em] \multirow{2}{*}{Row} & \multirow{2}{*}{Method}  & \multirow{2}{*}{LDR Supervision} & \multicolumn{2}{c|}{Training} & \multicolumn{3}{c|}{HDR} & Inference          \\
&        &     & Time (min)   & Memory (MB)    & PSNR$\uparrow$    & SSIM$\uparrow$    & LPIPS$\downarrow$ & speed (fps)     \\
\midrule[0.1em] 
1  & NeRF \cite{nerf} & $\times$  & 2147   & 14160 & 12.58  & 0.372  & 0.793  & 0.02            \\
2  & 3DGS \cite{3dgs} & $\times$  & \underline{27}     & \textbf{4008}  & 23.39   & 0.699  & 0.502   & \underline{155}      \\
3  & 3DGS-D4          & $\times$  & \textbf{26}     & \underline{4726}   & 26.69  & 0.741  & \underline{0.317}     & 138     \\
4  & 3DGS-D5          & $\times$           & 28     & 4774   & \underline{26.82}  & \underline{0.742}     & \underline{0.317}   & 141             \\
5  & Mono-HDR-GS \cite{mono-hdr-3d}  & \checkmark   & 156    & 16127  & 16.63     & 0.668   & 0.337  & 61     \\
6  & Mono-HDR-NeRF \cite{mono-hdr-3d} & \checkmark  & 570    & 19165  & 14.49     & 0.493   & 0.842 & 0.02    \\
7  & \textbf{LCD-GS (Ours)} & $\times$  & \textbf{26}  & 4879 & \textbf{27.74} & \textbf{0.749} & \textbf{0.295}  & \textbf{214}   \\
\bottomrule[0.15em]
\end{tabular}
} 
\label{tab:res_real}
\end{minipage}

\begin{minipage}{\linewidth}
\centering
\includegraphics[width=1.0\linewidth]{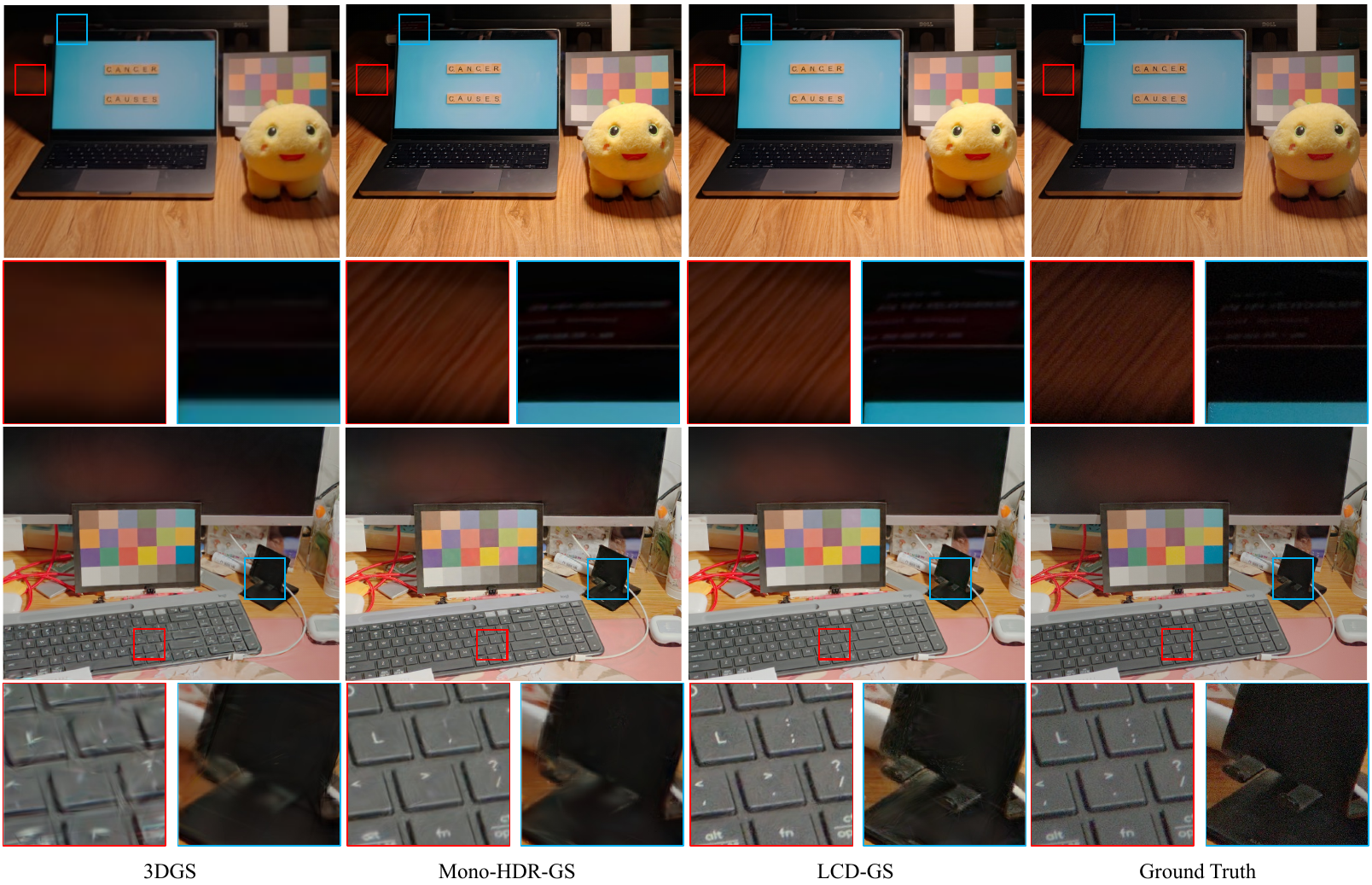}
\vspace{-1.0em}
\captionof{figure}{Visual comparisons on the iHDR-4S dataset.} \label{fig:real_cmp}
\end{minipage}
\vspace{-2.0em}
\end{figure}

Tab. \ref{tab:res_syn} and Tab. \ref{tab:res_real} present the results on the HDR-Syn-8S dataset and our proposed iHDR-4S dataset, respectively. We highlight the following key points:

\textbf{(I) Superior HDR reconstruction quality.} Our model significantly outperforms all LDR-focused competitors \cite{3dgs,nerf,scaffold-gs}. Although we replace the bounded sigmoid output with softplus to allow unbounded HDR prediction, NeRF \cite{nerf} still struggles under direct HDR supervision, likely due to the difficulty of optimizing high-dynamic-range radiance and geometry jointly through an implicit MLP representation.
Relative to vanilla 3DGS \citep{3dgs}, LCD-GS achieves a substantial +9.50 dB PSNR gain on the synthetic dataset and +4.35 PSNR gain on the real-world dataset, attributable to our luminance-chromaticity decomposition that resolves the fundamental limitation of entangled SH representations in HDR regimes. While increasing the maximum SH degree (\textit{e.g.}, to 5th-degree) marginally improves 3DGS's HDR performance, such approaches introduce significant computational overhead (41\% slower inference) and may exhibit pronounced overfitting artifacts, as shown in Fig. \ref{fig:sh3vs4} and discussed in Sec. \ref{sec:method:elevating}. In contrast, our approach preserves the explicit, optimization-friendly structure of 3DGS while decoupling luminance from chromaticity, enabling seamless integration into any SH-based 3DGS framework without architectural overhaul.

\textbf{(II) Advantages of native HDR representation.} 
LCD-GS consistently surpasses HDR-NeRF \cite{hdrnerf} and other HDR NVS methods that depend on multi-exposure LDR inputs \cite{hdr-gs, gausshdr} across all evaluated metrics. This comparison highlights the benefit of using native HDR observations: when such data are available, they provide direct radiometric cues and avoid the exposure calibration, view alignment, and tone-mapping ambiguities introduced by LDR-to-HDR pipelines \cite{hdrnerf,hdr-gs,gausshdr}. Rather than reconstructing HDR radiance through an intermediate LDR domain, LCD-GS optimizes the 3D representation directly in the HDR observation space, reducing error propagation and eliminating the need for additional tone-mapping or fusion modules. These results indicate that native HDR supervision offers a simpler and more reliable route to HDR scene reconstruction, enabling more accurate radiance estimation and more stable color reconstruction.

\textbf{(III) Computational efficiency.} LCD-GS maintains inference speed, training time, and memory usage on par with vanilla 3DGS across both synthetic
and real-world datasets, demonstrating that the proposed luminance–chromaticity decomposition incurs negligible computational overhead.
Compared to prior HDR methods, LCD-GS is 2,000$\times$ faster than HDR-NeRF \citep{hdrnerf} and nearly 2$\times$ faster than HDR-GS \cite{hdr-gs}. This efficiency stems from our explicit parametric representation: all HDR attributes are optimized as direct Gaussian primitive parameters rather than latent network outputs, eliminating the need for auxiliary neural networks while preserving 3DGS's real-time rendering capability and extending its dynamic range coverage.

\subsection{Qualitative results}
As shown in Fig. \ref{fig:syn_cmp}, both 3DGS and HDR-GS struggle to recover structures in dark regions, even though HDR-GS is trained with both HDR supervision and multi-exposure LDR guidance. In contrast, LCD-GS reconstructs these areas with clearer and more complete details. Similarly, Fig. \ref{fig:real_cmp} shows that both 3DGS \cite{3dgs} and Mono-HDR-GS \cite{mono-hdr-3d} tend to produce floaters or blurred structures in low-luminance regions, whereas LCD-GS yields more faithful scene representations while preserving finer details.
\begin{figure}
\center
\includegraphics[width=\linewidth]{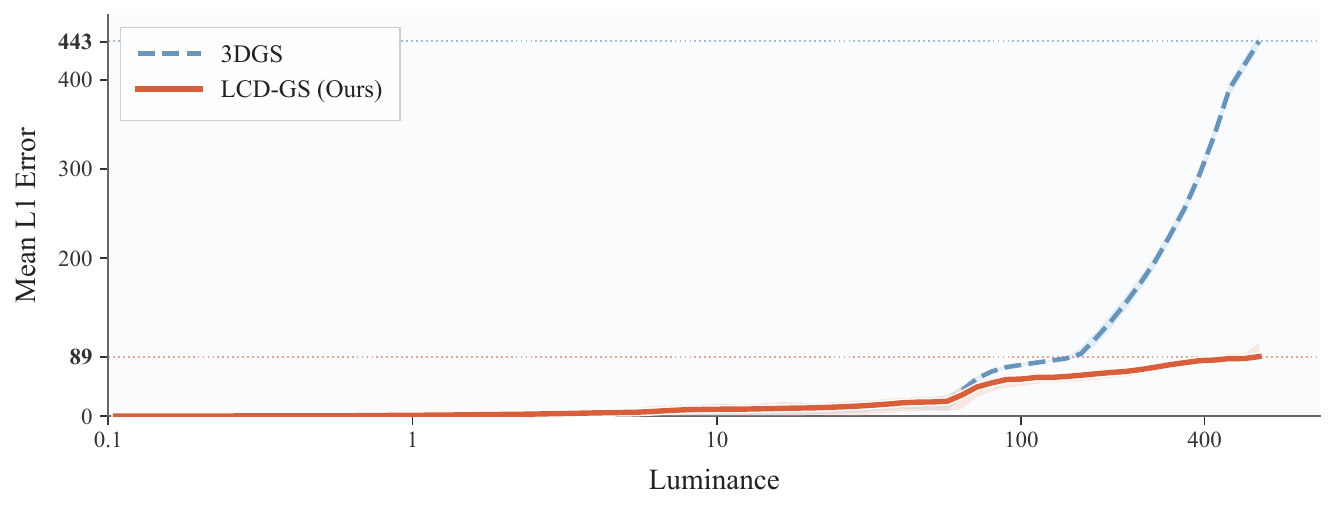}
\vspace{-2em}
\caption{Luminance-error profile. Mean $\ell_1$ error across varying luminance. While vanilla 3DGS suffers from exponentially escalating errors in high-intensity HDR regions, LCD-GS maintains robust and consistent accuracy.}
\vspace{-2.0em}
\label{fig:intensity_curves}
\end{figure}
Fig. \ref{fig:intensity_curves} further further analyzes the reconstruction error as a function of ground-truth luminance on the desk scene, with additional comparisons provided in Appendix \ref{sec:app:ablation}. The error-luminance profile shows that vanilla 3DGS maintains relatively small errors in low- and mid-luminance regions, but its mean $\ell_1$ error increases rapidly once the ground-truth luminance enters the high-intensity HDR range. This indicates that the color representation used in vanilla 3DGS becomes increasingly insufficient for modeling large radiance variations, especially when the apparent radiance changes sharply with viewing direction. In contrast, LCD-GS exhibits a much slower error growth and remains substantially more accurate in high-luminance regions. This improvement stems from our luminance--chromaticity decomposition: the luminance component explicitly models the radiance magnitude, while the chromaticity component focuses on bounded view-dependent color variation. As a result, LCD-GS provides more faithful HDR reconstruction in bright regions, which is critical for applications requiringaccurate lighting representation, such as AR/VR~\citep{van2022novel}.

\section{Conclusion}
We introduced LCD-GS, a lightweight HDR 3D Gaussian Splatting framework that learns directly from native HDR observations without relying on multi-exposure LDR supervision. Our analysis shows that the SH-based color representation in vanilla 3DGS entangles radiance magnitude and chromaticity, limiting its ability to model the large dynamic range and sharp radiance variations in HDR scenes. LCD-GS addresses this issue through a luminance-chromaticity decomposition, where a positive luminance scale captures the dominant HDR radiance magnitude and a bounded view-dependent chromaticity term models normalized color variation. This simple modification preserves the original 3DGS pipeline while substantially improving HDR reconstruction quality. 
Experiments on both HDR-Syn-8S and iHDR-4S demonstrate that LCD-GS achieves superior reconstruction fidelity and dynamic-range preservation compared with LDR-oriented baselines and existing HDR NVS methods, while maintaining efficient training and real-time rendering. Its explicit luminance component further provides a simple handle for luminance editing naturally.
{ 
    \small 
    \bibliographystyle{unsrtnat} 
    \bibliography{neurips_2026} 
}

\appendix
\newpage
\section{Appendix}
\label{sec:app}

\subsection{Datasets}
\label{sec:app:datasets} 
\begin{figure}[htbp]
\centering
\includegraphics[width=\linewidth]{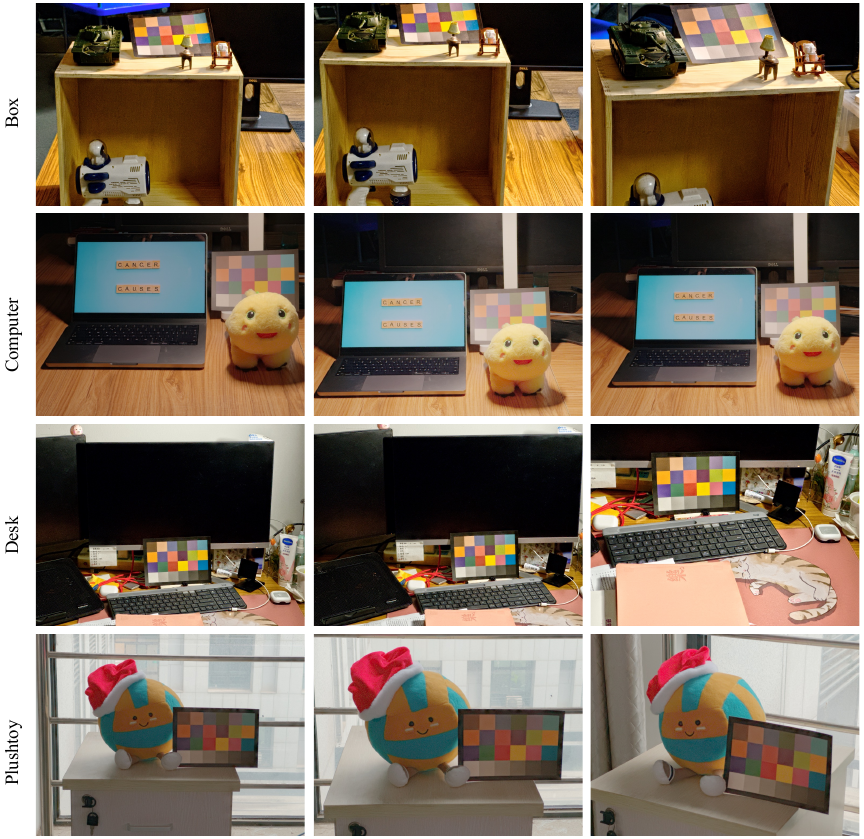}
\caption{Dataset composition for four scenes of iHDR-4S. Each row corresponds to a distinct scene with HDR images tone-mapped by Photomatix Pro for display.}
\label{fig:app:examples}
\end{figure}

This section provides a detailed description of the iHDR-4S
dataset introduced in this work.
\begin{wraptable}
{r}{0.45\textwidth}
\vspace{-1.2em}
\centering
\caption{Statistics of iHDR-4S.}
\vspace{0.5em}
\renewcommand{\arraystretch}{0.9}
\resizebox{0.45\textwidth}{!}{%
\begin{tabular}{c|cccc}
  \toprule[0.15em] Scenes & Training & Testing & Total & Resolution                        \\
  \midrule[0.1em] Box     & 39       & 5       & 44    & \multirow{4}{*}{4032$\times$3024} \\
  Computer                & 48       & 6       & 54    &                                   \\
  Desk                    & 49       & 7       & 56    &                                   \\
  Plushtoy                & 59       & 8       & 67    &                                   \\
  \bottomrule[0.15em]
\end{tabular}%
}
\label{tab:app:dataset}
\vspace{-0.5em}
\end{wraptable}
The dataset is captured using a single iPhone 14 Pro mounted on a tripod to ensure accuracy and stability, with the ProCam application\footnote{https://www.procamapp.com/} in ProRAW format. It is important to clarify that iPhone ProRAW images are genuinely high dynamic range; unlike standard 8-bit LDR formats (\textit{e.g.}, JPEG or HEIC) that suffer from irretrievable illumination clipping, ProRAW natively outputs 12-bit linear DNG files. By circumventing non-linear tone mapping and quantization, these 12-bit linear captures accurately preserve the wide dynamic range of physical scene radiance.
We then convert the HDR images into their LDR counterparts, as COLMAP \cite{colmap} does not support HDR images as input for point cloud reconstruction from multi-view images. Furthermore, Tab.~\ref{tab:app:dataset} summarizes the dataset statistics and Fig.~\ref{fig:app:examples} shows some example HDR images of different scenes.



\subsection{Additional related work}
\label{sec:app:add_related}
\textbf{Illumination editing} in neural 3D representations is commonly studied through relighting and inverse rendering \citep{hubert2025editing}. NeRF-based methods \citep{srinivasan2021nerv, boss2021nerd, rudnev2022nerf, sun2023sol} typically decompose scene appearance into geometry, reflectance, visibility, and illumination, enabling novel lighting synthesis by modifying the estimated lighting or material components. 
Recent works \citep{ye2023intrinsicnerf, radl2024laenerf} further extend this direction to intrinsic decomposition and lighting-aware neural fields, aiming to separate reflectance and shading for controllable appearance editing. 
More recently, illumination-aware editing has also been explored in 3D Gaussian Splatting \citep{liang2024gs, liang2025gus, sun2025svg, shi2025gir, du2025gs, cui2025luminance}. These approaches are powerful for general relighting, but they often introduce extra networks \citep{liang2024gs, shi2025gir}, physically based rendering assumptions \citep{wu20253d}, or dedicated illumination estimation modules \citep{huang2026emissive}. 

Unlike these relighting-oriented methods, LCD-GS does not aim to recover a full physical decomposition of lighting, material, and visibility, nor does it target arbitrary light-source editing. Instead, it provides a lightweight and direct form of \emph{luminance-based HDR scene editing}. By explicitly decoupling each Gaussian primitive into a view-independent luminance scalar and a normalized chromaticity term, LCD-GS enables global or local manipulation of radiance magnitude while keeping chromaticity unchanged. This makes our editing operation particularly suitable for HDR-specific adjustments, such as enhancing shadowed regions (see Fig. \ref{fig:app:lum_edit_shadow}) or compressing overly bright areas (see Fig. \ref{fig:app:lum_edit}), without introducing undesired hue shifts.

\subsection{Additional visualization results} 
\label{sec:app:exps} 
\begin{figure}[tbp]
\centering
\includegraphics[width=\linewidth]{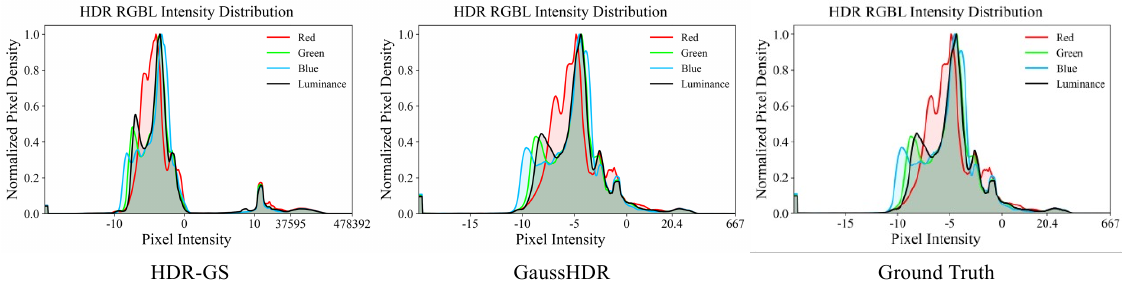}
\vspace{-1.5em}
\caption{Additional RGB and luminance curves of the rendered image on the HDR-Syn-8S dataset (desk scene), comparing HDR-GS \citep{hdr-gs} and GaussHDR \citep{gausshdr} with the ground truth.}
\label{fig:app:app_exps}
\vspace{-1em}
\end{figure}

This section provides further visualization of the HDR NVS performance of GaussHDR \cite{gausshdr} and HDR-GS \cite{hdr-gs}, as shown in Fig. \ref{fig:app:app_exps}.
Consistent with the observations in Fig. \ref{fig:intensity_curves}, it is evident that while both HDR-GS and GaussHDR achieve highly competitive quantitative metrics (\textit{e.g.}, PSNR, SSIM, and LPIPS) on the HDR-Syn-8S dataset (see Tab. \ref{tab:res_syn}), they still fundamentally fail to restore the full dynamic luminance range of the scene.

\begin{figure}
\centering
\includegraphics[width=\linewidth]{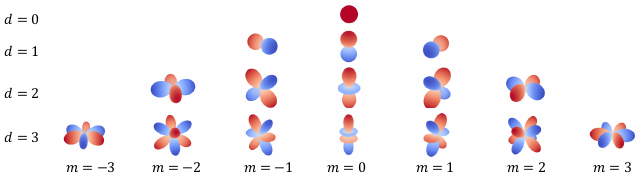}
\caption{Basis functions of spherical harmonics used in 3DGS, indexed by degree $d$ and order $m$. The shape of each function tells the spatial distributions of these harmonic modes, with different colors representing the variation in the function's values (red/blue indicating positive/negative values).}
\label{fig:3d-sh}
\vspace{-1.5em}
\end{figure}

\begin{figure}[tbp]
\centering
\includegraphics[width=\linewidth]{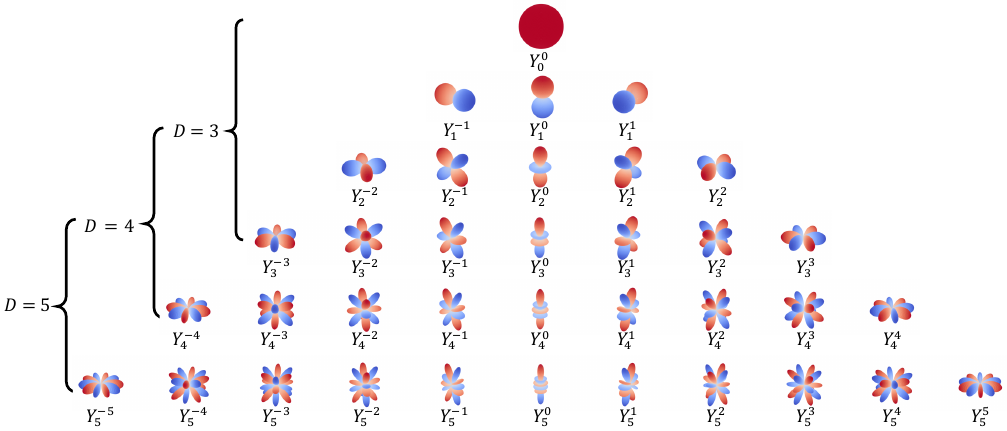}
\caption{Spherical harmonic basis functions: maximum degrees $D=3,4,5$ correspond to $(D+1)^2 = 16, 25, 36$ coefficients, respectively.}
\label{fig:app:shs_vis}
\vspace{-1.5em}
\end{figure}

\subsection{Ablation study}
\label{sec:app:ablation}
We evaluate three luminance activation functions (\textit{i.e.}, softplus, exponential and ReLU) on both the HDR-Syn-8S and iHDR-4S datasets and observe that softplus consistently outperforms the other two. Its smooth, non-saturating gradients facilitate stable convergence and accurate luminance estimation. In contrast, ReLU's thresholding behavior for negative inputs results in gradient truncation, which impedes optimization. While the exponential function preserves gradients, its unbounded nature causes slight numerical instabilities. As a result, softplus proves to be the most robust choice for HDR luminance modeling, as demonstrated in Tab. \ref{tab:app:ablation}.

\begin{table}[tbp]
\centering
\caption{Ablation studies of different luminance activation functions on both HDR-Syn-8S and iHDR-4S datasets. The \textbf{best} and \underline{second-best} results on each dataset are highlighted.}
\vspace{0.3em}
\resizebox{\textwidth}{!}{
\begin{tabular}{l|c|cc|ccc|c}
\toprule[0.15em]
\multirow{2}{*}{Activation} & \multirow{2}{*}{Dataset} & \multicolumn{2}{|c|}{Training} & \multicolumn{3}{c|}{HDR}   & Inference  \\
& & Time (min)    & Memory (MB) & PSNR $\uparrow$ & SSIM $\uparrow$  & LPIPS $\downarrow$ & speed (fps) \\
\midrule[0.1em]
Exponential  & \multirow{3}{*}{HDR-Syn-8S} & \textbf{11} & 2094 & \underline{41.14} & \underline{0.980} & \underline{0.007} & \textbf{250}  \\
ReLU &                             & \textbf{11} & \textbf{1334}  & 3.77   & 0.001 & 0.800 & \underline{237}         \\
Softplus    &     & \underline{14}  & \underline{1641}   & \textbf{42.69}  & \textbf{0.989} & \textbf{0.004} & 227  \\
\midrule[0.1em]
Exponential & \multirow{3}{*}{iHDR-4S}    & \underline{23} & 4927 & \underline{27.56}  & \underline{0.746} & \underline{0.303} & \underline{257} \\
ReLU &                             & \textbf{15}  & \textbf{3478}   & 3.32   & 0.001 & 0.665 & \textbf{274}   \\
Softplus    &  & 26  & \underline{4879}   & \textbf{27.74}  & \textbf{0.749} & \textbf{0.295} & 214    \\
\bottomrule[0.15em]
\end{tabular}
}
\label{tab:app:ablation}
\vspace{-1.0em}
\end{table}

\begin{figure}[tbp]
\vspace{-1em}
\centering
\includegraphics[width=\linewidth]{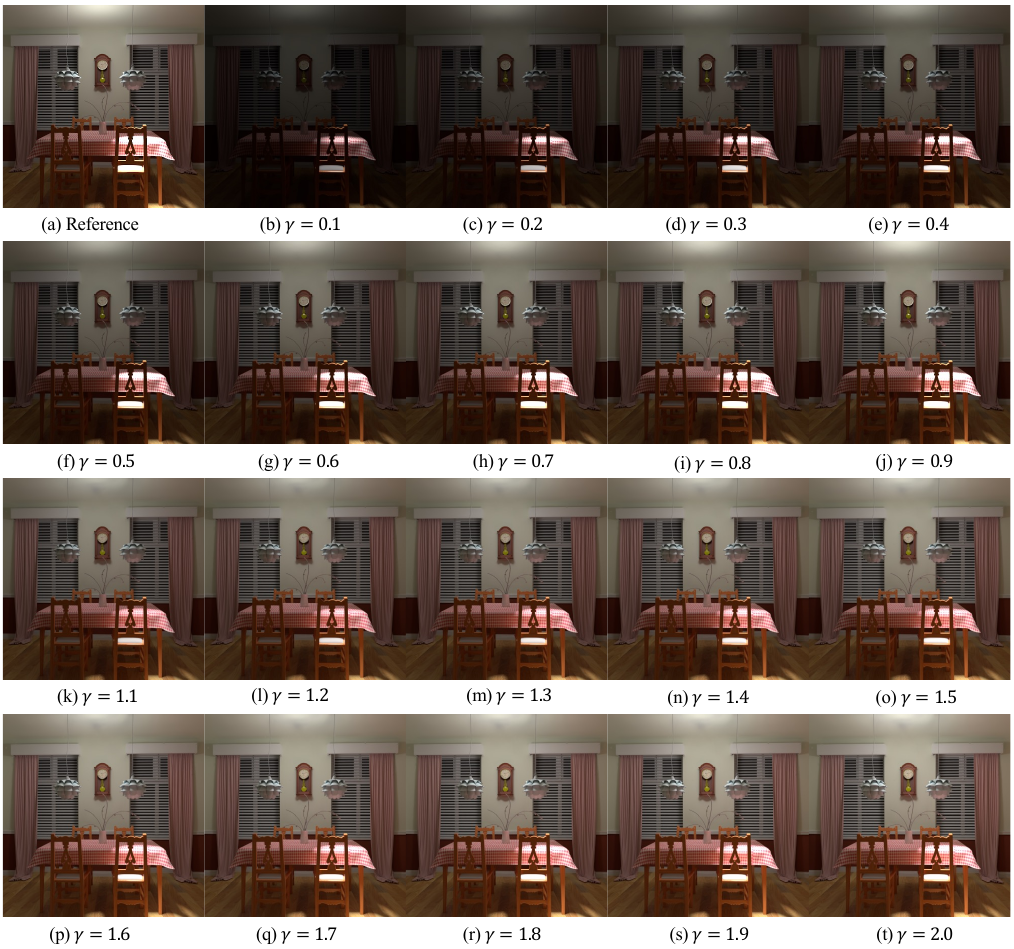}
\caption{
Global luminance editing with LCD-GS.
(a) Reference image.
(b)-(t) are obtained by scaling the learned luminance values of all Gaussian primitives with different $\gamma$ while keeping chromaticity, opacity, and geometry unchanged.
}
\label{fig:app:global_lum_edit}
\vspace{-1.5em}
\end{figure}

\begin{figure}[tbp]
\vspace{-2.em}
\centering
\includegraphics[width=\linewidth]{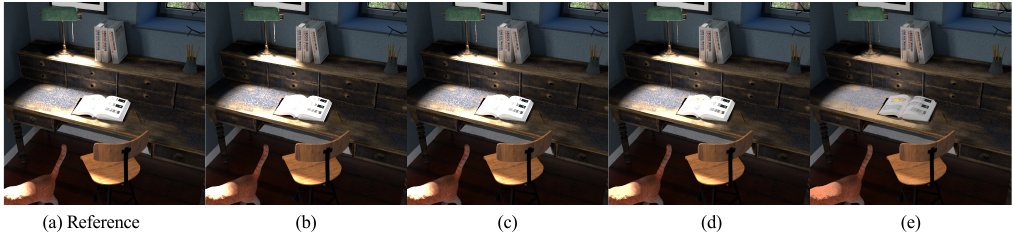}
\caption{Local luminance editing enabled by LCD-GS. (a) Reference image. (b)-(e) Highlight suppression by scaling the luminance of a subset of Gaussian primitives.}
\label{fig:app:lum_edit}
\end{figure}

\begin{figure}[tbp]
\vspace{-1em}
\centering
\includegraphics[width=\linewidth]{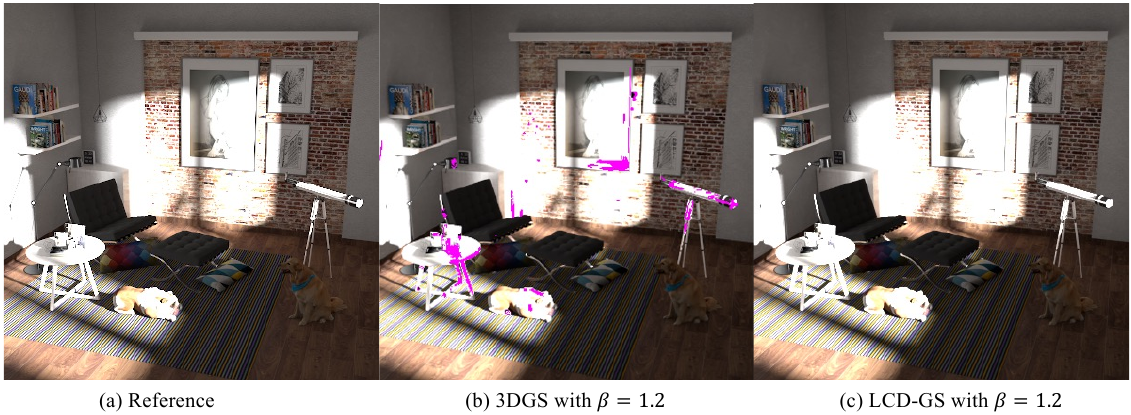}
\caption{
Comparison of global luminance editing between 3DGS and LCD-GS.
(a) Reference image.
(b) 3DGS result obtained by applying $\boldsymbol{c}'=\boldsymbol{c}^{\beta}$ with $\beta=1.2$ to the SH-predicted RGB colors of all Gaussian primitives.
(c) LCD-GS result obtained by applying $L'_{\mathrm{global}}=L^{\beta}$ with $\beta=1.2$ to the learned luminance values while keeping chromaticity, opacity, and geometry unchanged.
}
\label{fig:app:global_lum_edit_vs_3dgs}
\vspace{-1.5em}
\end{figure}

\subsection{Luminance editing}
\label{sec:app:scene_edition} 
This section provides additional analysis and examples of the luminance-based editing capability enabled by LCD-GS. Different from physically based relighting or inverse rendering methods, our goal is not to recover explicit light sources, materials, visibility, or illumination transport. Instead, LCD-GS offers a simple and direct interface for HDR radiance-magnitude editing by manipulating the learned luminance scalar of each Gaussian primitive while keeping its normalized chromaticity unchanged. This formulation supports both global luminance adjustment, where all primitives are transformed by a shared mapping function, and local luminance editing, where only selected primitives are modified before alpha compositing.

\subsubsection{Global luminance editing}
The explicit separation of luminance and chromaticity simplifies global luminance editing into a direct transformation. Under this framework, global exposure control can be viewed as a linear instantiation of the mapping function $f$ (other mappings, such as exponential functions, are also feasible; here we use a linear function as an illustrative example). In contrast to vanilla 3DGS \citep{3dgs}, which suffers from luminance-color entanglement, LCD-GS adjusts the scene's luminance by applying a scaling factor $\gamma \in \mathbb{R}^+$ to the luminance $L$:
\begin{equation}
L_{global}' = \gamma \cdot L,
\end{equation}
where $\gamma > 1$ simulates an increase in camera exposure or light source intensity, while $\gamma < 1$ produces a dimming effect, as shown in Fig. \ref{fig:app:global_lum_edit}.

The primary advantage of LCD-GS lies in its ability to preserve chromatic consistency during nonlinear luminance editing. In vanilla 3DGS, luminance manipulation must be applied directly to the SH-predicted RGB colors, since luminance and chromaticity are entangled in a single color representation. When a nonlinear transformation, such as $\boldsymbol{c}'=\boldsymbol{c}^{\beta}$, is applied to these RGB values, the relative ratios among the RGB channels are generally changed, leading to unintended artifacts, as shown in Fig. \ref{fig:app:global_lum_edit_vs_3dgs}b. In contrast, LCD-GS applies the nonlinear mapping only to the explicit luminance component, \textit{e.g.}, $L'=L^{\beta}$, while keeping the normalized chromaticity term unchanged. Therefore, the chromaticity remains invariant under global luminance remapping, enabling more stable luminance editing and better color consistency, as shown in Fig. \ref{fig:app:global_lum_edit_vs_3dgs}c.

\subsubsection{Local luminance editing}
As shown in Fig. \ref{fig:app:lum_edit} and Fig. \ref{fig:app:lum_edit_shadow}, LCD-GS also supports selective luminance editing at the Gaussian-primitive level. Specifically, we modify the learned luminance values of selected primitives during rendering while keeping chromaticity, opacity, and geometry unchanged. This enables targeted HDR luminance manipulation, such as suppressing highlights on the book in Fig. \ref{fig:app:lum_edit}b-e and enhancing low-luminance regions ($L \leq \tau$ where $\tau$ is a predefined threshold) on the floor in Fig. \ref{fig:app:lum_edit_shadow}b-e. Unlike global exposure adjustment, such local edits cannot be achieved by simply scaling the rendered image. Instead, they are performed before alpha compositing through the explicit luminance component, providing a simple and controllable interface for post-reconstruction radiance editing without introducing additional tone-mapping or color-correction modules.

\subsection{Limitations and broader impact}
\label{sec:app:limitation} 
The main limitations of this work are twofold. First, LCD-GS is more memory-demanding than vanilla 3DGS and requires slightly longer training time, particularly in complex scenes, which may hinder deployment on devices with limited memory budgets. Second, LCD-GS still depends on SfM-based Gaussian primitive initialization, making it sensitive to the quality of the initial sparse points and camera poses. In challenging scenarios where SfM becomes unreliable, the reconstruction quality may accordingly deteriorate.

The proposed method may benefit applications in immersive media, robotics, and digital scene preservation by improving 3D reconstruction under high dynamic range luminance. However, like other neural rendering techniques, it may also be misused for realistic scene replication, unauthorized environment capture, or privacy-invasive content generation.

\end{document}